# An interval-based aggregation approach based on bagging and Interval Agreement Approach in ensemble learning

Mansoureh Maadi[a], Uwe Aickelin[b], Hadi Akbarzadeh Khorshidi[c]
School of Computing and Information Systems
the University of Melbourne
Melbourne, Australia
mmaadi@student.unimelb.edu.au [a], uwe.aickelin@unimelb.edu.au [b], hadi.khorshidi@unimelb.edu.au [c]

*Abstract*—The main aim in ensemble learning is using multiple classifiers' outputs rather than one classifier output to aggregate them for more accurate classification. Generating an ensemble classifier generally is composed of three steps: selecting the base classifier, applying a sampling strategy to generate different individual classifiers and aggregation the classifiers' outputs. This paper focuses on the classifiers' outputs aggregation step and presents a new interval-based aggregation modeling using bagging resampling approach and Interval Agreement Approach (IAA) in ensemble learning. IAA is an interesting and practical aggregation approach in decision making which was introduced to combine decision makers' opinions when they present their opinions by intervals. In this paper, in addition to presenting a new aggregation approach in ensemble learning, we design some experiments to encourage researchers to use interval modeling in ensemble learning because it preserves more uncertainty and this leads to more accurate classification. For this purpose, we compared the results of implementing the proposed method to the majority vote, as the most commonly used and successful aggregation function in the literature, for 10 medical data sets. The results show the better performance of the interval modeling and the proposed interval-based aggregation approach in binary classification when it comes to ensemble learning. The Bayesian signed-rank test confirms the competency of our proposed approach.

## I. Introduction

Ensemble learning in machine learning operates based on the group decision making idea with the aim of learning from data using multiple classifiers than a single classifier. This process results better classification by decreasing the classifiers' variance and improving classification robustness and accuracy [1]. The application of ensemble learning in different real-world problems has been seen in various areas including image recognition, data mining, bioinformatics, information retrieval, etc. [1]. To create an ensemble classifier, there are some subtasks that should be considered: (1) selecting the base classifier to generate simple classifiers (2): choosing a sampling strategy for repeatedly sampling the training data and training the classifiers with them (3) aggregation the outputs of simple classifiers for a final output. To design an ensemble classifier, despite selecting the right base classifier, generating different sampling data sets and implementing a powerful aggregation strategy to aggregate the classifiers' outputs is necessary for more robust and accurate classification. In the literature, several sampling methods have been suggested which the main ones consist of Bagging [2], Dagging [3], Random Forest [4] and Boosting [5]. Also, to aggregate the simple classifiers' outputs, different aggregation strategies have been used which the most important and commonly used method is the majority vote to make final decision [6]. In this paper, we focus on the subtasks 2 and 3 of generating an ensemble classifier and propose a new interval-based aggregation approach based on bagging method and a new aggregation strategy named Interval Agreement Approach (IAA) for more accurate and robust classification. The IAA method is a method to model interval-valued data to fuzzy sets so that the loss of information be minimized and was proposed by Wagner et al. in 2015 [7].

IAA was proposed to aggregate the experts' opinions while their opinions are presented as intervals. In this paper, we use bagging strategy to make uncertainty intervals for each classifier. Considering classifiers as decision makers, we use IAA as a new aggregation function in ensemble learning to aggregate the uncertainty intervals. We believe interval modeling followed by interval aggregation functions can help capturing and storing uncertainty of simple classifiers' outputs to improve their performance for classification. So, the first question of this paper can be: does the proposed interval-based aggregation approach improve the accuracy of the classification? We design experiments to address this question by comparing the results of the proposed approach with the majority vote as the most commonly used aggregation function in the literature.

When it comes to decision tree as a base classifier in ensemble learning, using class probability decreases the accuracy of the classification [8], [9]. So, decision tree works generally based on class label rather than class probability. Majority vote as an efficient aggregation function has been used in ensemble learning when the outputs of the classifiers are labels not probabilities. The best application of the majority vote and decision tree has been seen in Random Forest (RF), as one of the best ensemble classifiers in the literature. In this paper, the second question is by designing interval modeling using class probabilities as the outputs of the decision trees and aggregating the uncertainty intervals implementing IAA, can we have better classification than when we use class labels as the outputs of the decision trees and the majority vote? Also, as the main problem of the majority vote is dependency to the base classifier, we designed an aggregation method using bagging and IAA to compensate this drawback to some extent by defining uncertainty intervals and using IAA so that the performance of classification can be improved in different binary classification problems.

Thus, the contribution of this paper is twofold. First, presenting a new interval modeling and aggregation approach in ensemble learning using classification probabilities, bagging strategy and IAA with the aim of capturing uncertainty in classification. Second, using class probability rather than class label as the output of the decision tree to show the power of interval modeling despite recommendation about using class label than class probability when it comes to decision tree classifier in the literature. The reminder of this paper is structured in the following manner. Section II summarizes the aggregation functions have been used in ensemble learning in the current literature. The proposed method is presented in section III. In Section IV, the designed experimental framework is described including description of the used data sets, set up information and evaluation methods. Section V provides a summary of results, comparisons and discussion following by section VI in which conclusion is presented.

## II. RELATED WORK

In the literature, several approaches have been presented to combine the outputs of classifiers in ensemble learning which the most important ones are Majority Vote, Weighted Voting, Naïve Bayes, Decision Templates and Stacking [10], [11]. Although, these methods have different strategies to aggregate the classifiers' outputs, the most presented approaches are based on an aggregation function to determine the final output label. The most common classical aggregation functions that have been applied in this area are averaging aggregation functions (means), weighted aggregation functions and majority vote [12]. Among these classical aggregation functions, majority vote has been broadly used in different ensemble methods as an efficient aggregation function. For example, Qamar et al. presented a majority vote based classifier ensemble technique using the ensemble of three classifiers of Naïve Bayes, Decision Trees and Support Vector Machines for web service classification [13]. Atallah and Al-Mousa presented a majority vote ensemble method for heart disease detection applying Random Forest, K-Nearest Neighbor, Logistic Regression and Stochastic Gradient Descent (SGD) classifiers [14]. Also, the application of the majority vote in ensemble learning can be seen in RF as a powerful ensemble learning approach.

In recent years, regarding the importance of fuzzy sets theory to deal with uncertainty especially in classification methods, another type of aggregation function presented in the literature is fuzzy integrals. Two fuzzy integrals named Choquet [15] and Sugeno [16] integrals have been proposed in the literature. These integrals aggregate the outputs of an ensemble of classifiers based on a fuzzy measure. In the literature, different ensemble classifiers based on fuzzy integrals have been presented which proposed different strategies to construct different fuzzy measures. As the review of different approaches to make fuzzy measures in ensemble learning is out of the scope of this paper, for more information, researchers referenced to papers [17] [18].

Interval-valued aggregation functions are a new class of aggregation functions that have been developed recently by several researchers and successfully applied in decision making problems [19], [20]. For example, in [19], different interval-valued aggregation functions are applied to design a fuzzy support system in medical diagnosis. The good performance of these aggregation functions in decision science specially to capture uncertainty can be a key factor to implement them in classification problems. The application of interval-valued aggregation functions is introduced recently in ensemble learning with the paper of Bentkowska et al. [21]. In [21], a new version of K-NN classifiers considering interval-valued aggregation functions is presented to deal with large numbers of missing values in data sets. In the proposed method, using K-NN classifiers and produced uncertainty intervals regarding missing values, different interval-valued aggregation functions which are based on the natural ordering of fuzzy sets are applied. Results on 10 data sets obtained from UCI machine learning repository show that the proposed approach has the better performance rather than arithmetic mean aggregation function. The future study of this paper is related to use of more effective and accurate integral-valued aggregation functions for better classification.

The main focus of this paper is applying an Interval-valued aggregation function named Interval Agreement Approach (IAA) to aggregate the classifiers' outputs in ensemble learning. IAA is a method to convert interval-valued data to fuzzy sets which is very practical in Computing with Words (CW) paradigm and linguistic summarization when experts are asked to determine their opinions over survey as intervals. The first method to convert a set of intervals to a fuzzy set is related to the paper of Liu and Mendel which introduced Interval Approach (IA) method in 2008 [22]. To overcome the drawback of the IA about considering some limitations regarding input data, Coupland et al. proposed an improved version of IA method named Enhanced Interval Approach (EIA) in 2010 [17]. Both methods presented in [22] and [17] were dependent to the data preprocessing, considered only limited types of fuzzy sets and didn't have ability to process uncertain intervals [7]. So, Miller et al. in 2012 introduced a new method to compensate the drawbacks of the previous methods by using interval-valued survey responses which considered intra- and inter person variability without any information lost during modeling to produce General Type-2 fuzzy sets from uncertain intervals [23]. Under synthetic and real-word numeric examples, Miller et al. concluded IA and their proposed method have their separate advantages and it is difficult to perform a comparison between them. Wagner et al. in 2015 presented a new method based on their method in [23] and IA known as Interval Agreement Approach (IAA) to convert interval-valued data to fuzzy sets [7]. The prominent characteristics of IAA is using advantages of both previous methods including consider minimal assumption about the distributions of the interval-valued data accompany with not relying on data preprocessing and outlier removal. So, this method can be very useful and practical when it comes to interval-valued data aggregation to get information.

Regarding the good performance of IAA to aggregate interval-valued data, we can see the applications or improvements of this method later in the literature. Navarro et al. in 2016 proposed a measure named Agreement Ratio based on IAA to show the overall aggregation among the experts given their survey responses in modelling words through fuzzy sets [24]. Application of this method on synthetic examples as well as a real-word dataset showed the proposed method can be a good analysis tool to recognize variation of perception of words in medical and patient relations specially when this variation results in misleading. Havens et al. in 2017 improved IAA method and proposed a new method named efficient Interval Agreement Approach (eIAA) [25]. eIAA provides a way to capture linguistical prototypes and computationally outperform IAA because this method has the potential of reducing necessary storage of fuzzy sets and computational complexity of fuzzy set operators. In 2020, Gunn et al. introduced a similarity measure for computing the similarity between two fuzzy sets resulted from IAA [26]. In this paper, they applied some features of the resulted fuzzy sets include centroids, perimeter, area, quartiles accompany with Principal Component Analysis (PCA) to construct this measure. To evaluate the proposed measure, an illustrative example related to comparison of films according to critics' interval scores is investigated.

According to the advantages of IAA to aggregate interval-valued data include minimal assumptions about the distributions of the input interval-valued data as well as not depending on data preprocessing, it seems this approach can be useful for better classification in ensemble learning specially in terms of capturing uncertainty. Also, as in IAA, both intra- and inter-person variability are modeled, this kind of modeling can be simulated in ensemble learning for better classification. Considering these kinds of variability, we are able to maintain uncertainty till the final class label be determined in classification. Also, as Interval modeling and aggregation is only applied recently in one paper, considering good performance of the proposed method in that paper for better classification, using IAA approach can be an important step to use interval modeling and interval based aggregation functions in ensemble learning. What if we could find an easy to use and practical aggregation strategy better than the majority vote?

III. METHOD

In this section, at first, we describe bagging strategy and IAA as the main bases of the proposed interval-based aggregation approach briefly. Afterward, steps of our method to determine the final class label in ensemble learning are explained.

*A. Bagging (Bootstrap Aggregation)*

Bagging is a resampling strategy which was proposed by Breiman in 1996 [2] with the aim of improving the accuracy and stability of the machine learning algorithms in classification and regression problems. This method is a simple as well as effective way to construct ensembles which help reduce variance and avoid overfitting in classification [18]. In this approach, to generate diversity among classifiers, a new data set is constructed for each classifier. This data set includes some instances elicited from the training dataset with replacement. Using this strategy, diversity among classifiers can be attained by training each classifier using different bootstrapped data set of the original training dataset. In the literature, this strategy is usually applied to decision tree methods. As said before, in ensemble learning, resampling is a method that can help a weak learner or classifier to achieve diversity for more accurate classification.

*B. Interval Agreement Approach (IAA)*

IAA is a method to generate fuzzy sets from interval-valued data and is presented and applied as a survey tool to analyze uncertainty in human responses when they represent the responses as intervals [25]. Consider an interval defined by $\bar{A} = [l_{\bar{A}}, r_{\bar{A}}]$, where $l_{\bar{A}}$ is the left endpoint and $r_{\bar{A}}$ shows the right endpoint. Regarding $\tilde{A} = \{\bar{A}_1, \ldots, \bar{A}_n\}$ as a set of intervals, in IAA a Type-1 Fuzzy Set (T1 FS) named $A$ can be created from intervals so that it represents the agreement among them. The membership function of $A$ is defined as follows:

$$\mu_A = \sum_{i=1}^{n} y_i \Big/ \left( \bigcup_{j_1=1}^{n-i+1} \bigcup_{j_2=j_1+1}^{n-i+2} \cdots \bigcup_{j_i=j_{i-1}+1}^{n} (\bar{A}_{j_1} \cap \ldots \cap \bar{A}_{j_i}) \right) \quad (1)$$

In (1), $y_i = i/n$ is the degree of membership. Also, "/" refers to the assignment of a given degree of membership not division. In IAA, the degree of membership is related to the number of intervals overlapped in a point. So, $y_i$ is equal to 1 only when all intervals are overlapped. To simplify (1), the membership function of A for a real number like $x$ can be rewrite as (2). According to (2), a degree of membership for $x$ is the frequency of which it appears within each interval divided by number of all intervals.

$$\mu_A(x) = \frac{\sum_{i=1}^{n} \mu_{\bar{A}_i}(x)}{n} \quad (2)$$

Where $\mu_{\bar{A}_i}(x) = \begin{cases} 1 & L_{\bar{A}_i} \leq x \leq r_{\bar{A}_i} \\ 0 & else \end{cases}$

For better description about how a T1 FS is generated by IAA, we describe it with an example. Consider three intervals of $\bar{A}_1 = [1, 4]$, $\bar{A}_2 = [2, 5]$ and $\bar{A}_3 = [3, 6]$ as input intervals for IAA. Regarding (2), $\mu_A(x)$ can be calculated for all real numbers $x$. According to the intervals, $\mu_A(x)$ has nonzero value for numbers in the range of [1,6]. For example, for x=1,3,5 $\mu_A$ is calculated as follows:

$$\mu_A(x=1) = \frac{1+0+0}{3} \quad \mu_A(x=3) = \frac{1+1+1}{3} \quad \mu_A(x=5) = \frac{0+1+1}{3} \quad (3)$$

After computing $\mu_A$ for all numbers in range [1,6], the final T1 FS can be depicted as Fig. 1. Regarding Fig. 1, T1 FS can be shown as a list of tuples on different regions of change over $\mu_A(x)$ as follows:

$T1\ FS=\{([1, 2), ^1/_3), ([2, 3), ^2/_3), ([3, 4], 1), ((4, 5], ^2/_3), ((5, 6], ^1/_3)\}$ (4)

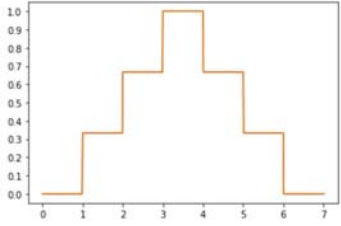

Fig. 1. Generating T1 FS from intervals $\bar{A}_1 = [1, 4]$, $\bar{A}_2 = [2, 5]$ and $\bar{A}_3 = [3, 6]$ using IAA.

### C. Proposed aggregation method based on bagging strategy and IAA

The aim of this paper is presenting a new aggregation approach based on bagging strategy and IAA in ensemble learning. As we said before, in ensemble learning, the first step is selecting a weak classifier as a base classifier. The second step is related to generating diversity among classifiers by choosing a sampling strategy to train the classifiers with sampled data. In the final step, the results provided by classifiers are aggregated to a single output label. In our proposed method, after selecting the base classifier, bagging strategy is applied as a resampling strategy to create uncertainty intervals. These intervals will be the input of IAA to aggregate them and create a T1 FS. Finally, the centroid of the generated T1 FS will be calculated to determine the final class label. In continue, the detailed steps of the proposed approach are described. Also, ensemble of classifiers aggregated by bagging strategy and IAA is shown in Fig. 2.

*Step1*. generation of different classifiers

In the first step, the base classifier is selected. In the literature, different classifiers like decision tree (DT), K-NN, SVM, etc. can be implemented as a base classifier to generate different classifiers. The output of this step is different classifiers to start step 2 of the proposed approach.

*Step2.* Generating uncertainty intervals using Bagging strategy

In this method, we capture the uncertainty of classifiers by using uncertainty intervals. In this step, Bagging strategy is used to generate uncertainty intervals. At first, a bootstrapped data set is generated from training dataset and all classifiers are trained on the bootstrapped data set. Considering n as the bootstrap size and m as the number of classifiers, this process will be repeated n times by generating n bootstrapped data sets. Considering a binary classification problem, in each repeat, the results of applying classifiers on test dataset is $P_i$ which shows the probability assigned by the classifier $i$ ($i = 1, \cdots, m$) to the main class.

At the end of using bagging strategy, $P_{ij}$ ($i = 1, \cdots, m$, $j = 1, \cdots, n$) are determined which show the probability assigned by classifier $i$ to the main class considering bootstrapped dataset j. Using $P_{ij}$ values, we can construct an uncertainty interval for each classifier $i$ using (5).

$$UI_i = [LQ_i, UQ_i] \quad (5)$$

In (5), $UI_i$ is an uncertainty interval which is determined for classifier i. As in statistical simulation, the Inter Quartile Range (IQR) is commonly used as robust measure of scale and generally is preferred to the total range, we suggest IQR to calculate uncertainty interval for each classifier. Also, as for each classifier $n$ bootstrapped data sets are generated and consequently $n$ probability values are calculated, to preserve useful knowledge as an interval for each classifier, ignoring outlier results seems necessary specially when the number of $n$ is high. In (5), Regarding $P_{ij}$ values for classifier $i$, $LQ_i$ is the first quartile of $P_{ij}$ ($j=1, ..., n$) and $UQ_i$ is the third quartile of $P_{ij}$ ($j=1, ..., n$). Using $m$ uncertainty intervals constructed for $m$ classifiers, the next step of the proposed method can be started.

*Step3.* Aggregation of uncertainty intervals using IAA

As said before, in this paper IAA is applied to generate a T1 FS. Considering each classifier as an expert and uncertainty intervals as experts' opinions, in this step we focus on capturing inter participant uncertainty through intervals in IAA. Using uncertainty intervals of the previous step and (1) or (2), we aggregate the intervals and the output of this step is a T1 FS named final T1 FS (FTFS) in this paper. Regarding (4), the FTFS is shown as (6).

$$R_i = ([R_{il}, R_{ir}], R_{ih}); \ FTFS = [R_1, R_2, ..., R_v] \quad (6)$$

In (6), FTFS is shown as a list of tuples while each tuple $R_i$ indicates different regions of change over the membership function. $l$ is the left point, $r$ is the right point and $h$ is the height or the membership function value of the tuple $R_i$. The next step of the proposed method is defuzzification of the calculated FTFS and determining the final class label.

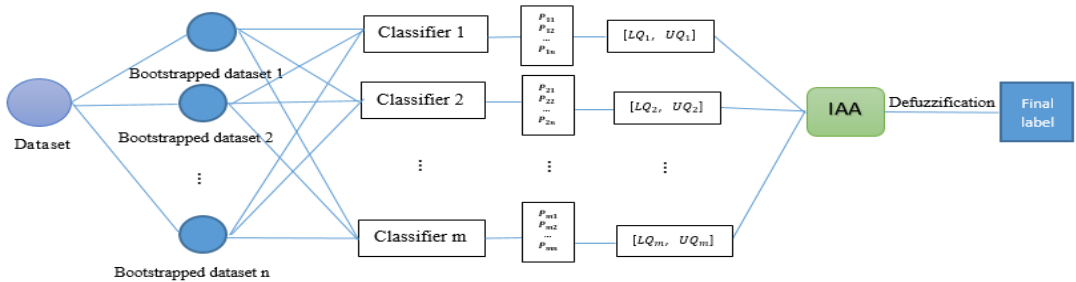

Fig. 2. Ensemble of classifiers aggregated based on bagging method and IAA

***Step4.*** Calculation of centroid for generated T1 FS and determining final class label

In this step, the centroid of the FTFS is calculated and used to determine the final class label. In this regards, using (7) the centroid of the FTFS is calculated.

$$centroid\ (FTFS) = \frac{\sum_{i=0}^{v}(R_{ih} \times R_{il}) + (R_{ih} \times R_{ir})}{\sum_{i=0}^{v} 2(R_{ih})} \quad (7)$$

In the proposed method, after calculation of the centroid, regarding the probability value of 0.5 in binary classification, the final class label of the data is determined. If the centroid is equal or upper than 0.5 the final class will be the main class (the class that we made our intervals based on that), otherwise another class would be the final class label.

## IV. EXPERIMENTAL FRAMEWORK

To assess the proposed method, we compare the performance of our suggested aggregation approach to the majority vote as the most commonly used aggregation methodology in the literature. Also, as decision tree is an important base classifier in ensemble learning, we design our experiments based on this classifier. We use class probability as the output of the decision tree in our proposed method to show the power of interval modeling in ensemble learning to capture uncertainty despite suggestions of using class label rather than class probability as the output of decision tree in the literature. Finally, to verify the good performance of our proposed aggregation approach compared to the majority vote, we implement a new nonparametric Bayesian version of the Wilcoxon signed-rank test on experiments.

### A. Data sets

In Table I, 10 data sets that have been utilized in our evaluation are listed with their characteristics. These data sets were collected from University of California at Irvin (UCI) Machine Learning Repository [27] and consist of different medical data sets. As disease diagnosis is a very important problem in binary classification, we implement medical data sets to evaluate our proposed method. Also, these data sets are different in terms of number of features, number of training and test samples, type of features, etc. So, they provide a comprehensive evaluation for our proposed method. We separate the training and test data sets for each disease following the strategy of [28]. The data sets organized in this paper can be downloaded from [29].

TABLE I. DESCRIPTION OF THE DATASETS USED IN THIS PAPER

| Data Set | NO. Of Features | Training Samples | Test Samples |
|---|---|---|---|
| Wisconsin Breast Cancer | 9 | 499 | 200 |
| Pima Indian Diabetes | 8 | 576 | 192 |
| Bupa | 6 | 200 | 145 |
| Hepatitis | 19 | 80 | 75 |
| Heart- Stat log | 13 | 180 | 90 |
| SpectF | 44 | 176 | 91 |
| SaHeart | 9 | 304 | 158 |
| Planning Relax | 12 | 120 | 62 |
| Parkinson | 22 | 130 | 65 |
| Hepatocellular Carcinoma (Hcc) | 49 | 110 | 55 |

### B. Setup for generating classifiers and evaluation

In our experiments, decision tree is applied as a base classifier for making different classifiers. To generate different classifiers, regarding *m* as the number of features, in each data set, we generate *m* decision trees by ignoring one feature in each decision tree. We have used decision tree classifier in sklearn package and the experiments are conducted using the Python programming language. To assess the quality of the proposed method, the accuracy and F-score rates are used in this paper. Each experiment is repeated 30 times and the mean of accuracy and the F-score are presented. Also, in each data set, the main class is specified for building uncertainty intervals which in medical science generally is the class related to getting infected to a disease.

## V. RESULTS AND DISCUSUIN

The results of accuracy and F-score for all data sets over different bootstrapped numbers (20, 50 and 100) for the proposed approach and the majority vote are shown in table II. The mean of accuracy and F-score values for the proposed method and the majority vote approach are calculated to show the performance of two approaches on medical data sets. To have a fair comparison, we use the same classifiers, the same data sets and the same bootstrapped data sets for two methods and the difference between the results are only due to different aggregation approaches performance.

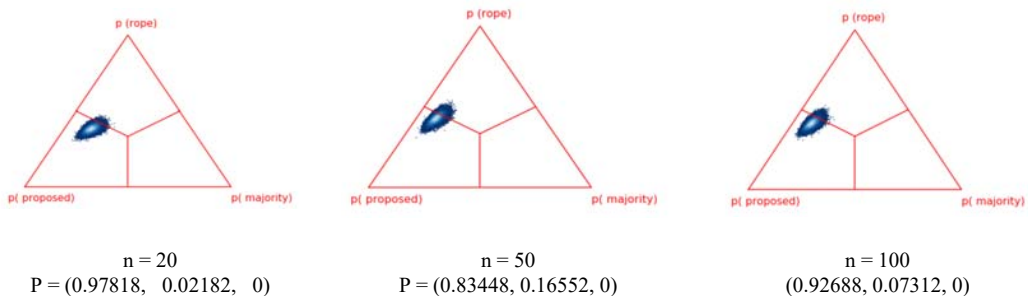

n = 20
P = (0.97818, 0.02182, 0)

n = 50
P = (0.83448, 0.16552, 0)

n = 100
(0.92688, 0.07312, 0)

Fig.3. Posterior plots and probabilities for the proposed approach vs. the majority vote in terms of accuracy on the data sets with n = 20,50 and 100 for the Bayesian signed-rank test

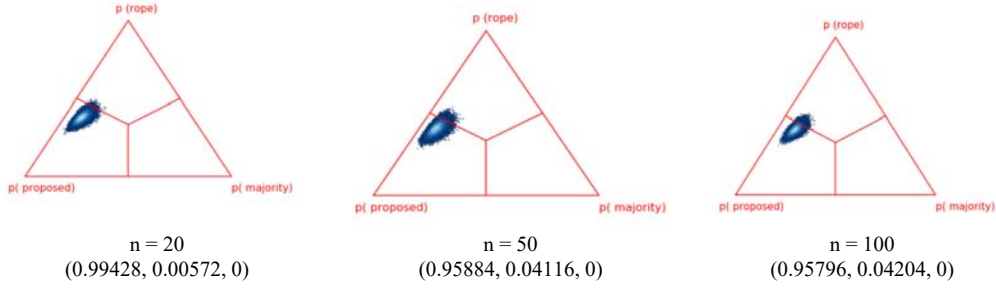

Fig.4 . Posterior plots and probabilities for the proposed approach vs. the majority vote in terms of F-score on the data sets with n = 20,50 and 100 for the Bayesian signed-rank test

In table II, for each medical data set, $n$ is the number of bootstrapped data sets generated from main training data set. In fact, $n$ is the number of a classifier's class probabilities used to produce uncertainty interval in the proposed approach. Regarding different values for $n$, we can see that the proposed method performs better than the majority vote approach in both terms of accuracy and F-score in all data sets.

To evaluate the significance of the superiority of the proposed method statistically, we use Bayesian signed-rank test which has all advantages of the Bayesian approach and is a new replacement for the frequentist signed and signed rank test in the literature [30]. The application of this statistical test returns to the comparison of two classification approaches over multiple data sets and its ability is answering questions about posterior probabilities which null hypothesis tests are not able to answer them. In fact, using this test, we can determine how high is the probability in which an approach is better than the other one more than 1% [30]. In Fig. 3, three posterior plots as the results of the Bayesian Signed-rank test as well as the related posterior probabilities regarding different numbers of $n$ are shown to compare the accuracy of the proposed approach and the majority vote. As is suggested in [30], to calculate the posteriors probabilities the prior parameter of the Dirichlet are considered $s = 0:5$ and $z_0 = 0$.

In the posterior plots, three regions are determined in a triangle and the clouds of points related to the samples from the posteriors are depicted. The region on the bottom left of triangle is associated with the case where the proposed aggregation method outperforms the majority vote. The top region is relevant to the situation where it is more probable that neither method is better. The bottom right region in triangle shows the status in which the majority vote is statistically better than the proposed method. Thus, more points in every bottom regions show that the approach related to that region has had statistically better performance.

In Fig. 4 the posterior plots and the probability values are shown to compare the proposed aggregation approach to the majority vote in terms of F-score. As can be seen, this figure validates that the associated probability of the better performance of the proposed method than the majority vote is higher than 0.957 regarding different values for $n$.

TABLE II. THE MEAN ACCURACY AND F-SCORE CALCULATED FOR THE PROPOSED AGGREGATION METHOD AND THE MAJORITY VOTE

| Data Sets | Number of bootstrapped data sets | Accuracy | | F- score | |
|---|---|---|---|---|---|
| | | **Proposed Approach** | majority vote | **Proposed Approach** | majority vote |
| Wisconsin Breast Cancer | n = 20 | **0.984** | 0.978 | **0.976** | 0.968 |
| | n = 50 | **0.984** | 0.979 | **0.977** | 0.969 |
| | n = 100 | **0.986** | 0.979 | **0.980** | 0.970 |
| Heart- Stat log | n = 20 | **0.824** | 0.813 | **0.774** | 0.765 |
| | n = 50 | **0.830** | 0.813 | **0.782** | 0.765 |
| | n = 100 | **0.820** | 0.810 | **0.776** | 0.763 |
| Bupa | n = 20 | **0.737** | 0.735 | **0.840** | 0.838 |
| | n = 50 | **0.736** | 0.734 | **0.839** | 0.838 |
| | n = 100 | **0.735** | **0.735** | **0.839** | 0.838 |
| Parkinson | n = 20 | **0.928** | 0.917 | **0.952** | 0.944 |
| | n = 50 | **0.927** | 0.918 | **0.951** | 0.945 |
| | n =100 | **0.925** | 0.920 | **0.949** | 0.946 |
| Hepatitis | n = 20 | **0.802** | 0.749 | **0.879** | 0.840 |
| | n = 50 | **0.802** | 0.748 | **0.879** | 0.840 |
| | n = 100 | **0.805** | 0.749 | **0.882** | 0.841 |
| Pima Indian Diabetes | n = 20 | **0.799** | 0.796 | **0.671** | 0.670 |
| | n = 50 | **0.798** | 0.796 | **0.668** | 0.667 |
| | n = 100 | **0.798** | 0.795 | **0.667** | 0.666 |
| SpectF | n = 20 | **0.829** | **0.829** | **0.900** | 0.899 |
| | n = 50 | **0.834** | 0.830 | **0.904** | 0.900 |
| | n = 100 | **0.828** | 0.822 | **0.900** | 0.896 |
| SaHeart | n = 20 | **0.743** | 0.725 | **0.519** | 0.464 |
| | n = 50 | **0.739** | 0.728 | **0.474** | 0.429 |
| | n = 100 | **0.752** | 0.735 | **0.504** | 0.444 |
| Planning Relax | n = 20 | **0.712** | 0.711 | **0.688** | 0.681 |
| | n = 50 | **0.713** | 0.709 | **0.688** | 0.681 |
| | n = 100 | **0.741** | 0.725 | **0.691** | 0.687 |
| Hepatocellular Carcinoma (Hcc) | n = 20 | **0.718** | 0.698 | **0.796** | 0.783 |
| | n = 50 | **0.720** | 0.693 | **0.795** | 0.782 |
| | n = 100 | **0.719** | 0.697 | **0.796** | 0.786 |

To answer the first question of this research, as described above, the proposed approach has better performance rather than the majority vote in terms of accuracy and F-score in all data sets. So, it seems the proposed interval modeling and aggregation approach is a competitive approach to improve the accuracy of the classification in ensemble learning. In our experiments, as the number of classifiers is equal to the number of features in each data set, regarding variety range of number of features in the data sets, we tested our proposed aggregation method considering different number of classifiers and find out that our aggregation method is not dependent on the number of classifiers to perform better than the majority vote in all data sets. Also, as we evaluate the proposed method for different values of $n$ as the number of bootstrapped data sets and consequently the number of classification probabilities to generate uncertainty interval for each classifier, according to the results, we can conclude that our proposed algorithm outperforms the majority vote regardless of whether $n$ is large or small. However, more experiments need in future to investigate different parameters involved in better performance of interval modeling rather than the commonly used aggregation functions.

In this paper, we use decision tree as the base classifier in ensemble learning and design our experiments based on the outputs of this classifier. Although, it is recommended in the literature to work with class label rather than class probability when it comes to use decision tree, regarding table II, we can see that by implementing interval modeling and class probability, the classification results are more accurate rather than when we use class labels and majority vote. So, to answer the second question of this paper, we can say that by using interval modeling and the proposed interval-based aggregation approach we have the chance to use class probability rather than class label as the output of decision tree and have better classification results. This shows the power of interval modeling and IAA in ensemble learning.

## VI. CONCLUSION

Interval modeling and interval aggregation approaches in decision making especially when experts are asked to determine their opinions over survey as intervals have been attracting interest of researches to propose different approaches for information fusion or information aggregation. These approaches have had outstanding achievements for better and more accurate decision making with the aim of capturing uncertainty in different research areas. IAA is an interesting and practical aggregation approach introduced to combine interval-valued data produced by experts' responses into fuzzy sets.

In ensemble learning, after generating classifiers, to aggregate the classifiers' outputs, different aggregation approaches have been proposed in the literature. As uncertainty is inevitable in classification and ensemble learning, we have proposed an interval modeling in ensemble learning to capture the uncertainty. For this purpose, we have used bagging to generate uncertainty intervals for each classifier and considering classifiers as decision makers implemented IAA to aggregate the uncertainty intervals to a fuzzy number. After that, by defuzzification, we have determined the class label of the data samples. We have evaluated our proposed method on ten medical data sets in binary classification and compared the results of the proposed approach to the majority vote as the most commonly used aggregation function in the literature. The results have shown the better performance of our proposed algorithm in terms of accuracy and F-score. To verify this outperformance, we have used a new nonparametric statistical test named the Bayesian signed-rank test.

Regarding our experiments, we have tested the advantage of interval modeling to capture uncertainty when we are using the class probability rather than class label as the output of decision tree. In spite of suggestions about emphasis on class label rather class probability when we are using decision tree in ensemble learning, interval modeling and IAA have been successful in classification in our experiments. The results of these experiments show that interval modeling and IAA using class probability outperforms the majority vote using class labels. The contributions of this paper can be summarized as follows:

- Introducing a new interval modeling to capture uncertainty for better classification in ensemble learning

- Using IAA as a recently introduced interval-valued aggregation function in ensemble learning

- Preserving useful and effective information for classifiers using bagging and IQR in ensemble learning

As there is only one paper [21] has been presented in the literature of ensemble learning, which uses interval modeling to handle missing values issue, there is a need for more research in this area. Both paper [21] and our study show that interval modeling can be practical and successful in ensemble learning. For the future works, based on the proposed approach some suggestions are presented as follows:

- Implementing other base classifiers like logistic regression, KNN, and so on to test the performance of the proposed aggregation approach

- Using other resampling techniques to generate uncertainty intervals to aggregate them by IAA

- Developing the proposed method for multiclass and multilabel classification problems.